%% file: Camera_ready.tex
\documentclass{article}

\usepackage{microtype}
\usepackage{graphicx}
\usepackage{subfigure}
\usepackage{booktabs} 

\usepackage{hyperref}

\usepackage[accepted]{icml2024}

\usepackage{amsmath}
\usepackage{amssymb}
\usepackage{mathtools}
\usepackage{amsthm}

\usepackage[capitalize,noabbrev]{cleveref}

\theoremstyle{plain}
\newtheorem{theorem}{Theorem}[section]
\newtheorem{proposition}[theorem]{Proposition}
\newtheorem{lemma}[theorem]{Lemma}

\theoremstyle{definition}

\theoremstyle{remark}

\DeclareMathOperator*{\argmax}{arg\,max}

\usepackage[textsize=tiny]{todonotes}

\usepackage{subfiles}

\icmltitlerunning{A Unified LP Framework for Offline Reward Learning}

\begin{document}

\twocolumn[
\icmltitle{A Unified Linear Programming Framework for Offline Reward Learning\\ from Human Demonstrations and Feedback}



\icmlsetsymbol{equal}{*}

\begin{icmlauthorlist}
\icmlauthor{Kihyun Kim}{equal,mit}
\icmlauthor{Jiawei Zhang}{equal,mit}
\icmlauthor{Asuman Ozdaglar}{mit}
\icmlauthor{Pablo Parrilo}{mit}
\end{icmlauthorlist}

\icmlaffiliation{mit}{Department of Electrical Engineering and Computer Science, Massachusetts Institute of Technology, Cambridge, USA}

\icmlcorrespondingauthor{Kihyun Kim}{kihyun@mit.edu}

\icmlkeywords{Machine Learning, ICML}

\vskip 0.3in
]



\printAffiliationsAndNotice{\icmlEqualContribution} 

\begin{abstract}
Inverse Reinforcement Learning (IRL) and Reinforcement Learning from Human Feedback (RLHF) are pivotal methodologies in reward learning, which involve inferring and shaping the underlying reward function of sequential decision-making problems based on observed human demonstrations and feedback.
Most prior work in reward learning has relied on prior knowledge or assumptions about decision or preference models, potentially leading to robustness issues.
In response, this paper introduces a novel linear programming (LP) framework tailored for offline reward learning.
Utilizing pre-collected trajectories without online exploration, this framework estimates a feasible reward set from the primal-dual optimality conditions of a suitably designed LP, and offers an optimality guarantee with provable sample efficiency.
Our LP framework also enables aligning the reward functions with human feedback, such as pairwise trajectory comparison data, while maintaining computational tractability and sample efficiency.
We demonstrate that our framework potentially achieves better performance compared to the conventional maximum likelihood estimation (MLE) approach through analytical examples and numerical experiments.

\end{abstract}

\subfile{sections/introduction}

\subfile{sections/section2}
\subfile{sections/section3}
\subfile{sections/section4}

\subfile{sections/experiment}

\section{Conclusion}
In this paper, we have introduced a novel LP framework designed for offline reward learning.
Our framework possesses several salient features, including $(i)$  tractability and sample efficiency with an optimality guarantee, $(ii)$ flexibility for extension due to its convex solution set, and $(iii)$ robustness against diverse decision models.
We believe our study opens a new avenue of research in the theories of offline reward learning.
In the future, we aim to extend our framework to broader datasets, including those involving arbitrary sampling policies in IRL and $K$-wise comparisons in RLHF.
Additionally, we plan to investigate the transferability of the estimated reward functions to similar environments.

\section*{Acknowledgments}
The authors would like to express their gratitude for the support that made this work possible. J. Zhang was supported by the MIT School of Engineering Postdoctoral Fellowship Program for Engineering Excellence. A. Ozdaglar acknowledges support from the AI Accelerator program. 
\section*{Impact Statement}
This paper aims to advance the understanding and efficiency of reward learning. We believe our work has the potential to benefit sequential decision-making systems in fields such as robotics, autonomous driving, and language models, though no immediate specific societal concerns are directly raised by our findings.

\bibliography{reference}
\bibliographystyle{icml2024}

\newpage
\appendix
\onecolumn
\subfile{sections/appendix}


\end{document}

%% file: sections/introduction.tex
\section{Introduction}
Reward learning involves inferring and shaping the underlying reward function from observed human demonstrations and feedback.
Inverse reinforcement learning (IRL) and reinforcement learning from human feedback (RLHF, also known as preference-based reinforcement learning) are key methodologies in reward learning, applied in various sequential decision-making tasks such as games~\citep{macglashan2017interactive, christiano2017deep, ibarz2018reward}, robotics~\citep{finn2016guided, brown2019extrapolating, shin2023benchmarks}, and language models~\citep{ziegler2019fine, stiennon2020learning, wu2021recursively, ouyang2022training, liu2023chain}.s
Particularly in the recent drastic development of large language models (LLMs), RLHF has played a crucial role in fine-tuning models to better align with human preferences~\citep{ouyang2022training}.
However, despite the notable empirical success of these algorithms, a significant gap remains in the theoretical analysis of IRL and RLHF, limiting us to guarantee their reliability.
This work aims to bridge this gap by proposing a novel theoretical framework for offline IRL and RLHF.

IRL aims to infer a reward function that aligns with an expert behavior from demonstrations~\citep{ng2000algorithms, abbeel2004apprenticeship}.
Typical IRL algorithms employ a bi-level optimization framework within the context of maximum likelihood estimation (MLE).
In this framework, the inner optimization evaluates the policy based on the current reward parameters, while the outer optimization updates these parameters to better match observed expert behavior.
These algorithms have been extensively explored in the literature~\citep{ziebart2008maximum, wulfmeier2015maximum, zhou2017infinite, zeng2022structural, zeng2023understanding}, and their convergence is studied in both online settings~\citep{zeng2022structural} and offline settings~\citep{zeng2023understanding}.

Building upon IRL, RLHF recovers a reward function from human feedback data, which is commonly given by pairwise or $K$-wise trajectory comparisons. 
Despite growing interest in RLHF, only a few algorithms provide optimality guarantees with sample complexity bound.
Most existing approaches adopt an MLE framework, assuming that human evaluators follow a presupposed preference model, such as the Bradley-Terry-Luce (BTL) model.
The reward parameters are then fine-tuned to maximize the log-likelihood of the collected offline preference data~\citep{christiano2017deep}.
Recently, \citep{zhu2023principled, zhan2023provable, li2023reinforcement} have proposed offline RLHF algorithms with optimality guarantees by adopting a pessimistic mechanism from offline reinforcement learning (RL) theory.
\citep{park2024principled} further suggested a personalized setting that learns multiple reward models for heterogeneous preferences.

While the MLE-based approach offers a solid theoretical framework for both IRL and RLHF, it comes with unavoidable limitations.
Particularly in IRL, bi-level optimization algorithms face computational challenges due to their nested-loop structures.
In addition, MLE-based algorithms rely on a specific decision-making or preference model they employ.
For example, the IRL algorithm proposed by~\citep{zeng2022structural} learns the reward function that aligns with an expert policy, which is assumed to be an optimal softmax policy.
Furthermore, RLHF algorithms (e.g. \citep{zhu2023principled, zhan2023provable}) assume a preference model for human evaluator, which might not fully capture the complex and diverse nature of real-world human preferences.
Consequently, their optimality guarantees might be compromised if there exists a  mismatch between actual human preferences and the model in use.

In response to these challenges inherent in MLE frameworks, recent research has shifted focus from estimating a single reward function (under a presupposed model) to recovering a \emph{feasible reward set}, a set of rewards where the expert policy is (near) optimal.
Notably, \citep{metelli2021provably, metelli2023towards} estimated the feasible reward set from finite-horizon Bellman equations and provided sample complexity bounds associated with estimation errors.
However, their algorithm requires a generative model of state transition probabilities.
\citep{lindner2022active} mitigated this requirement by adopting an efficient exploration policy for sampling trajectories, though it remains in an online setting.
More recently, and concurrent with our work, \citep{zhao2023inverse} introduced the first offline algorithm with a theoretical guarantee.
They introduce a pessimistic mechanism to address the issue of non-uniform data coverage, penalizing state-action pairs with low visitation frequency. 
Nevertheless, these penalty functions are nonlinear and non-convex, resulting in a non-convex reward set.
This could limit flexibility for applications, especially when selecting a specific reward function within the set.

In line with this evolving perspective, we aim to obtain a \emph{convex estimate} of a feasible reward set in an offline setting.
To achieve this, we leverage recent advancements in the Linear Programming (LP) framework within the domain of offline RL.
A fundamental challenge in offline RL is the non-uniform coverage of offline data ~\citep{fujimoto2019off, kumar2020conservative}.
To address this issue, recent algorithms have employed a pessimistic mechanism that conservatively selects the value function or model within an uncertainty set~\citep{liu2020provably, jin2021pessimism, rashidinejad2021bridging, xie2021bellman,uehara2021pessimistic, chen2022offline}.
However, these pessimistic approaches often introduce non-convex optimization problems, which can be intractable.
In the latest research, a series of works~\citep{zhan2022offline, rashidinejad2022optimal, ozdaglar2023revisiting} have introduced LP-based methods that relax data coverage assumptions and provide tractable algorithms suitable for function approximation by introducing convex formulations.

Given the success of LP-based approaches in offline RL, investigating how it could address non-convexity and non-uniform data coverage issues in offline IRL presents a promising research direction.
One notable advantage of LP is its flexibility in addressing intrinsic challenges in reward learning, such as avoiding undesirable degenerate solutions like $r=\mathbf{0}$.
We demonstrate that a polyhedral estimate of the feasible reward set, provided by LP, offers efficient ways to identify a non-degenerate reward function.
For example, it allows to select a reward function that maximizes the reward gap between the expert policy and suboptimal policies (e.g., uniform policy) over the solution set.
We also highlight LP's suitability for function approximation, primarily due to its linear structure, which can further reduce the solution set and computational complexity.
Furthermore, the LP framework enables the integration of extra information.
As a notable example, we show that RLHF data can be incorporated by simply adding linear constraints, maintaining computational tractability and sample efficiency.

Our main contributions can be summarized as follows:
\begin{itemize}
    \item We present an LP formulation for offline IRL that directly estimates the feasible reward set by the primal-dual optimality conditions in an empirical LP formulation of Markov decision process (MDP) (Section~\ref{sec:IRLformulation}).
    \item In Theorem~\ref{thm:optimality}, the optimality of the estimated reward set is provided such that any reward function within this set ensures the expert policy is $\tilde{O}(\sqrt{|S||A|/N})$-suboptimal, under appropriate data coverage assumption.
    \item In offline RLHF, we align reward functions with pairwise trajectory comparison data using linear constraints (Section~\ref{sec:RLHFformulation}).
    In Theorem~\ref{thm:rlhf}, we provide the generalization guarantee of the estimated reward function for unseen trajectory pairs.
    \item We address the potential degeneracy issue in reward learning (Section~\ref{sec:IRLdegeneracy}) and propose a unified framework that effectively combine IRL and RLHF to mitigate the degeneracy (Section~\ref{sec:integration}).
    \item The proposed LP algorithm and the MLE algorithms in the literature are compared through numerical experiments (Section~\ref{sec:experiment}).
    We also provide an analytical example in offline RLHF, where MLE algorithm fails, while our LP approach succeeds to identify the optimal policy (Appendix~\ref{app:mle})
\end{itemize}
Additional related literature is discussed in Appendix~\ref{app:relatedwork}.

%% file: sections/section2.tex
\section{Preliminaries}

\subsection{Markov Decision Process (MDP)}
We first revisit the standard notations for a tabular infinite-horizon discounted Markov decision process (MDP).
An MDP $\mathcal{M}$ is represented as a tuple $\mathcal{M} = (S, A, P, \gamma, \mu_0, r)$, where $S$ and $A$ represent finite state and action spaces, $P: (S, A) \mapsto \Delta(S)$ denotes the transition probability function, $\gamma \in (0, 1)$ represents the discount factor, and $\mu_0 \in \Delta(S) $ is the initial state distribution.
The reward function $r(s,a): S \times A \mapsto [-1, 1]$ indicates the reward received for taking the action $a$ in state $s$.

The primary objective in MDP is to identify a stochastic policy $\pi:S \mapsto \Delta(A)$ that maximizes the expected cumulative reward: $\mathbb{E}^\pi [\sum_{h=0}^\infty \gamma^h r(s_h, a_h) | s_0 \sim \mu_0]$.
We define $v^\pi(s)$ as the expected total discounted reward received when initiating from state $s$ and following $\pi$, such that $v^\pi(s) := \mathbb{E}^\pi\left[\sum_{h=0}^\infty \gamma^h r(s_h, a_h) \mid s_0 = s\right]$.
Then, the optimal policy $\pi^*$ maximizing the expected reward and its corresponding value function $v^* := v^{\pi^*}$ are related by the Bellman equation
\begin{equation}
v^*(s) = \max_{a \in A} \left\{ r(s,a) + \gamma \sum_{s' \in S} P(s' | s, a) v^*(s') \right\},
\end{equation}
which holds for any $s \in S$.
It is well-established that $v^*$ can be determined by solving the following linear programming (LP) as outlined in~\citep{puterman2014markov}:
\begin{equation}
\min_{v \in \mathbb{R}^{|S|}} \;  (1-\gamma)\mu_0^\top v \quad \text{s.t.} \quad M^\top v - r \geq 0.
\end{equation}
The matrix $M \in \mathbb{R}^{|S|^2|A|}$ is defined as  $M(s', (s,a)) := \mathbf{1}_{\{s' = s\}} - \gamma P(s' | s,a)$, where $\mathbf{1}_{\{s' = s\}}$ denotes the indicator function for the case $\{s'=s\}$.
Throughout the paper, we treat the above LP as the primal LP, and $v$ as the primal optimization variable.
Then, the dual LP is expressed as
\begin{equation}
\max_{d\in \mathbb{R}^{|S||A|}} \;  r^\top d \quad \text{s.t.} \quad  M d = (1-\gamma)\mu_0 , \quad  d \geq 0.
\end{equation}
The dual variable $d$, often interpreted as an \emph{occupancy measure} or a discounted state-action visitation frequency in RL literature, is related to a policy $\pi$ by
\begin{equation}
d^\pi (s, a) = (1-\gamma) \sum_{h=0}^\infty \gamma^h P^\pi_{\mu_0} (s_h = s, a_h = a).
\end{equation}
Here, $P^\pi_{\mu_0} (s_h = s, a_h = a)$ represents the probability of $(s_h, a_h) = (s, a)$, given $s_0 \sim \mu_0$ and $a_{h'} \sim \pi(s_{h'})$ for all $h' \geq 0$.
The dependence on $\gamma$ and $\mu_0$ is omitted in the notation $d^\pi$ for simplicity.
A more detailed relationship between $\pi$ and $d^\pi$ can be found in~\citep{puterman2014markov}.

\subsection{Offline IRL: Learning from Expert Trajectories}
While RL learns a policy to maximize rewards in a given environment, IRL aims to infer the underlying reward function that drives observed behavior.
In the standard IRL setting, a single expert agent collects trajectory (roll-out) samples, and the reward function is recovered from these samples.
We denote the true expert policy and corresponding occupancy measure as $\pi_e$ and $d_e$, respectively, where $d_e$ is defined as $d_e := d^{\pi_e}$.
Our objective is to learn a reward function $r$ such that the occupancy measure $d_e$ is (near) optimal, utilizing the offline dataset gathered from the expert policy $\pi_e$. 

In offline setting, the true values of the expert policy $\pi_e$ and the transition probability $P$ are unknown.
Instead, we have access to a static, pre-collected dataset $\mathcal{D}_\text{IRL}$ composed of $N$ independent and identically distributed (i.i.d.) trajectory samples:
\begin{equation}
\mathcal{D}_\text{IRL} = \{\tau^n = (s_0^{n}, a_0^{n}, s_1^{n}, \ldots, s_{H-1}^{n}, a_{H-1}^{n}, s_H^{n} )\}_{n=1}^N.
\end{equation}
Note that the sampling distribution is fully determined by $\mu_0$, $P$, and $\pi_e$.
Let $N_h(s,a)$ and $N_h(s,a,s')$ be the counts of $n$ satisfying $(s_h^n, a_h^n) = (s,a)$ and $(s_h^n, a_h^n, s_{h+1}^n) = (s, a, s')$ in the dataset, respectively.
Using these counts, we estimate the occupancy measure $d_e$ as follows:
\begin{equation}
    \hat{d}_e(s,a) =  (1-\gamma) \frac{1}{N} \sum_{h=0}^{H-1} \gamma^h N_h (s,a) \quad \forall (s,a) \in S \times A.
\end{equation}
Using this empirical estimate $\hat{d}_e$, we aim to develop an LP formulation that identifies a reward function which ensures the optimality of $d_e$ with an acceptable level of error.


\subsection{Offline RLHF: Learning from Pairwise Trajectory Comparisons}\label{sec:formulation:feedback}
We extend our LP framework to address the offline RLHF problem. Our primary objective is to derive a reward function that is aligned with pairwise trajectory comparison data, provided by human evaluators.
We denote each comparison query as $q_n$, with the index $n$ ranging from 1 to $N_q$.
Each query comprises a pair of trajectories, such that $\tau^{n, 1} = (s_0^{n, 1}, a_0^{n, 1}, \ldots, s_{H-1}^{n, 1}, a_{H-1}^{n, 1}, s_H^{n, 1})$ and $\tau^{n, 2} = (s_0^{n, 2}, a_0^{n, 2}, \ldots, s_{H-1}^{n, 2}, a_{H-1}^{n, 2}, s_H^{n, 2})$.
We assume $\tau^{n, 1}$ and $\tau^{n, 2}$ are sampled i.i.d. according to the sampling distribution $\mu_\text{HF}$.
In each query, a human evaluator is presented with both trajectories and asked to select the one they prefer. 
We denote the event where trajectory $\tau^{n,1}$ is preferred over $\tau^{n,2}$ by the variable $y^n = 1$, and conversely, $y^n = 2$ indicates the event where $\tau^{n,2}$ is favored over $\tau^{n,1}$.
The human feedback dataset is then represented by $\mathcal{D}_\text{HF} = \{ (\tau^{n, 1}, \tau^{n, 2}, y^n) \}_{n=1}^{N_q}$.

Given the dataset $\mathcal{D}_\text{HF}$, we design an LP formulation to identify a reward function $r$ that aligns well with $\mathcal{D}_\text{HF}$.
Notably, this LP approach is purely data-driven, relying solely on the observed comparisons without assuming any specific preference model associated with human evaluators.
This aspect distinguishes it from previous MLE algorithms for offline RLHF.
The detailed comparison will be elaborated in a later section of the paper.

%% file: sections/section3.tex
\section{Offline Inverse Reinforcement Learning}
\subsection{LP Formulation of Offline IRL}\label{sec:IRLformulation}
Recall the dual LP formulation of MDP presented in the previous section:
\begin{equation}\label{eq:dual}
\begin{split}
\max_{d\in \mathbb{R}^{|S||A|}} \;  r^\top d \quad \text{s.t.} \quad  M d = (1-\gamma)\mu_0 , \quad  d \geq 0.
\end{split}
\end{equation}
IRL aims to find a feasible reward function $r$ such that the expert occupancy measure $d_e$ is an optimal solution to the above dual LP.
As the feasible reward function is not unique, we define the feasible reward set as follows:
\begin{equation}
    \mathcal{R} = \{r \in [-1, 1]^{|S||A|} \mid d_e \text{ is optimal to \eqref{eq:dual}}\}
\end{equation}
In the offline setting, recovering the ground-truth $\mathcal{R}$ is challenging since we only have access to the empirical estimate of $d_e$ and the transition matrix $P$.
Consequently, our goal is to recover an estimate of $\mathcal{R}$ in which $d_e$ is a near-optimal solution to~\eqref{eq:dual} for any $r$ in this estimated set.

\paragraph{Marginal importance sampling.}
The primary challenge in offline RL and IRL is the non-uniform coverage of the offline dataset.
To address this issue, we adopt the marginal importance sampling (MIS) framework in the literature~\citep{nachum2019algaedice, lee2021optidice}, which considers the scaled version of LP.
First, we define the optimization variable $w_d \in \mathbb{R}^{|S||A|}$ as
\begin{equation}\label{eq:w}
w _d(s,a) := \begin{cases}\frac{d(s,a)}{d_e(s,a)} \quad &\text{if} \quad d_e(s,a) > 0, \\ 0 \quad &\text{if} \quad d_e(s,a) = 0. \end{cases}
\end{equation}
$w_d$ is a scaled dual variable, which represents the ratio between the target $d$ and the expert $d_e$.
The expert optimization variable is denoted by $w_e := w_{d_e}$, which satisfies  $w_e (s,a) = \mathbf{1}_{\{d_e(s,a) > 0\}}$ for all $(s, a) \in S \times A$ by the definition of $w_d$.
Note that our algorithm will not require information about which state-action pairs have zero visitation frequency under $\pi_e$ (i.e., $d_e(s,a)=0$), since it will automatically set the reward to zero, i.e. $r(s,a)=0$, if $\hat{d}_e(s,a)=0$.
 
Next, we define $u\in \mathbb{R}^{|S||A|}$ and $K \in \mathbb{R}^{|S|^2|A|}$ as
\begin{equation}
\begin{split}
u(s,a) & := r(s,a) d_e (s,a),\\
K(s', (s,a)) & := d_e (s,a) \mathbf{1}_{\{s= s'\}} - \gamma d'_{e} (s, a, s'),
\end{split}
\end{equation}
where $d'_e (s, a, s') := d_e(s,a) P(s'|s,a)$ for any $(s,a,s')$.
In this MIS framework, $u$ and $K$ correspond to $r$ and $P$, respectively.
The following lemma shows this relationship clearly (see Lemma 1 in~\citep{ozdaglar2023revisiting} for the proof).
\begin{lemma}\label{lem:substitution}
 $r^\top d = u^\top w_d$ and $Md = Kw_d$ hold for any $d\in \mathbb{R}^{|S||A|}$.
\end{lemma}

\paragraph{Empirical LP formulation.}
By Lemma~\ref{lem:substitution}, the dual LP can be written with $u$, $w$, and $K$ as follows:
\begin{equation}
\max_{w \in \mathbb{R}^{|S||A|}} \; u^\top w \quad \text{s.t.} \quad  K w =  (1-\gamma)\mu_0,\quad w \geq 0.
\end{equation}
We omit the subscript $d$ in $w_d$ for ease of notation.
In the above LP formulation, our objective is to identify the set of $u$ for which $w_e$ is optimal.
However, in the offline setting, the true values of $K$ and $d_e$ remain unknown.
Therefore, our goal shifts towards constructing an empirical version of this LP.
We first define the empirical estimate of $u$ as $u_\mathcal{D}(s,a) := r(s,a) \hat{d}_e(s,a)$ for all $(s,a) \in S \times A$, and replace the objective function $u^\top w$ with $u_\mathcal{D}^\top w$.
Next, we introduce $K_\mathcal{D} \in \mathbb{R}^{|S|^2|A|}$, an empirical estimate of $K$, defined as:
\begin{equation}
K_\mathcal{D}(s', (s,a)) := \hat{d}_e(s,a) \mathbf{1}_{\{s = s'\}} - \gamma \hat{d}'_e(s, a, s'),
\end{equation}
where for any $(s,a, s') \in S \times A \times S$,
\begin{equation}
    \hat{d}'_e(s,a, s') := (1-\gamma) \frac{1}{N} \sum_{h=0}^{H-1} \gamma^h N_h(s,a, s').
\end{equation}
However, directly substituting the empirical estimate $K_\mathcal{D}$ for $K$ in the equality constraint $Kw = (1-\gamma)\mu_0$ can be problematic, as it may cause the target variable $w_e$ being infeasible.
 Therefore, we opt to relax the equality constraint to an inequality constraint.

Let $X = [ x_1, \cdots, x_{N_X} ] \in \mathbb{R}^{|S|\times N_x} $ be a coefficient matrix for the relaxation, where $\Vert x_i \Vert_\infty \leq 1 $ for all $i \in \{1, \ldots, N_x\}$.
Let $\epsilon_x \in \mathbb{R}^{N_x}$ be a parameter that controls the level of relaxation.
Then, we replace the equality constraint $Kw = (1-\gamma) \mu_0$ with the relaxed inequality constraint $X^\top (K_\mathcal{D} w - (1-\gamma)\mu_0) \leq \epsilon_x$.
One applicable choice of the coefficient matrix $X$ would be a matrix that contains all $2^{|S|}$ binary (sign) vectors $[\pm 1, \pm 1, \ldots, \pm 1]$ in its columns.
Then, the inequality constraint is equivalent to the $\mathcal{L}^1$ norm constraint, i.e. $\Vert K_\mathcal{D} w - (1-\gamma)\mu_0 \Vert_1 \leq \epsilon_x$.

With this relaxation, the empirical version of the dual LP can be expressed as
\begin{equation}\label{eq:dual_offline}
\begin{split}
\max_{w \in \mathbb{R}^{|S||A|}} \quad &   u_\mathcal{D}^\top w \\
\text{s.t.} \quad & X^\top (K_\mathcal{D} w - (1-\gamma)\mu_0) \leq \epsilon_x, \quad w \geq 0.
\end{split}
\end{equation}
Additionally, the dual of \eqref{eq:dual_offline} can be expressed as
\begin{equation}\label{eq:primal_offline}
\begin{split}
\min_{v  \in \mathbb{R}^{N_x} } \quad &   (1-\gamma)\mu_0^\top X v + \epsilon_x^\top v\\
\text{s.t.} \quad & K_\mathcal{D}^\top X v \geq u_\mathcal{D}, \quad v \geq 0,
\end{split}
\end{equation}
where $v$ is an optimization variable.

\paragraph{Feasible reward set estimation.}
Under the empirical LP formulations, our goal is to estimate the set of $u$ such that $w_e$ is (near) optimal to~\eqref{eq:dual_offline}.
Consider the primal-dual optimality conditions of $(v, w)$ under a reward function $u$:
\begin{equation}\label{eq:kkt}
\begin{split}
    &\text{(Primal feasibility)}: K_\mathcal{D}^\top X v \geq u, \; v \geq 0,\\
    &\text{(Dual feasibility)}: X^\top (K_\mathcal{D} w - (1-\gamma)\mu_0) \leq \epsilon_x, \; w \geq 0,\\
    &\text{(Zero duality gap)}: (1-\gamma)\mu_0^\top X v + \epsilon_x^\top v = u^\top w.
\end{split}
\end{equation}
$w$ is dual-optimal under $u$ if and only if the above optimality conditions hold with some $v$. 

Consequently, the feasible reward set can be estimated by identifying $(u, v)$ pairs for which $(u, v, w_e)$ satisfies \eqref{eq:kkt}.
Here, we further relax the zero duality gap condition with the slack parameter $\epsilon_g \geq 0$ for two reasons. First, the true reward might not satisfy this equality constraint due to errors in empirical estimation of $d_e$ and $K$.
Second, we are also interested in the reward such that $\pi_e$ is near-optimal.
Therefore, we consider the following polyhedron as an estimate of the feasible reward set:
\begin{equation}\label{eq:solution}
\begin{split}
\hat{\mathcal{R}}_\text{IRL}(\epsilon_g) :=  \{(u, v) \mid \underbrace{(1-\gamma)\mu_0^\top X v + \epsilon_x^\top v   -  u^\top \mathbf{1} \leq \epsilon_g}_{(i)},\\
\underbrace{K_\mathcal{D}^\top X v \geq u, \; v \geq 0}_{(ii)}, \; \underbrace{ -\hat{d}_e \leq u \leq \hat{d}_e}_{(iii)} \}.
\end{split}
\end{equation}
The constraint $(i)$ denotes the upper bound on the duality gap, where $\epsilon_g$ is used as the parameter.
$(ii)$ represents primal feasibility condition and $(iii)$ bounds the reward $r$ to the range $[-1, 1]$.
The vector $\mathbf{1}$ (vector of all ones) is used instead of $w_e$ in $(i)$, since $u^\top \mathbf{1} = u^\top w_e$ holds by the definition of $u$ and $w_e$.
Note that the dual feasibility condition is not required in $\hat{\mathcal{R}}_\text{IRL}(\epsilon_g)$ because it is a condition for the constant value $w_e$.

\subsection{Optimality Guarantee for Offline IRL}\label{sec:IRLoptimality}
In this section, we analyze the statistical error involved in the estimate $\hat{\mathcal{R}}_\text{IRL}$ of the feasible reward set $\mathcal{R}$.
Before presenting the main results, we address the data coverage issue in our setting.
In offline RL, distribution mismatch between the target policy and the behavior policy causes the inaccurate policy evaluation, and the concentrability-type assumption is required for an optimality guarantee.
In offline IRL, since the behavior policy is identical to the expert policy, the reward estimation can be inaccurate for state-actions pairs where the occupancy measure $d_e(s,a)$ is small.
We address this issue by defining the confident set of occupancy measures.

\paragraph{Confidence set.}
The confidence set if defined as the intersection of a set of valid occupancy measure (under MDP $\mathcal{M}$) and an $\mathcal{L}^\infty$ norm ball with radius $B$:
\begin{equation}
D_B :=  \left\{ d \in \mathbb{R}_+^{|S||A|} \mid M d = (1-\gamma)\mu_0, \; \Vert w_d \Vert_\infty \leq B \right\}.
\end{equation}
The radius $B\geq1$ is a parameter that controls the conservativeness of the algorithm. 

The set $D_B$ includes all possible occupancy measures $d$ if $B \geq d_\text{min}^{-1}$, where
\begin{equation}
d_\text{min} := \min_{(s,a) \in S \times A:\;d_e(s,a) \neq 0}d_e(s,a),    
\end{equation}
since $w_d(s,a) \leq d_\text{min}^{-1}$ for any $d \in \Delta(S\times A)$ and $(s,a) \in S \times A$ by the definition of $w_d$~\eqref{eq:w}.
In this case, optimality over the set $D_B$ implies global optimality.

It is worth highlighting that setting $B = d_\text{min}^{-1}$ yields results comparable to those in recent works~\citep{metelli2023towards, zhao2023inverse}.
The error bounds in these works depend on the constant $\pi_\text{min}^{-1}$, defined as
\begin{equation}
\pi_\text{min} := \min_{(s,a) \in S \times A:\;\pi_e(a | s) \neq 0} \pi_e(a | s).
\end{equation}
Under a fixed $\mu_0$, the value of $B = d_\text{min}^{-1}$ is upper bounded by $(\text{constant}) \times \pi_\text{min}^{-1}$, by the following inequality:
\begin{equation}
\begin{split}
    d_\text{min} &= d_e (s',a') = \pi_e(a' | s') \sum_{a \in A} d_e (s',a)\\
    &\geq (1 -\gamma) \mu_0(s') \pi_\text{min},    
\end{split}
\end{equation}
where $(s', a') \in S \times A$ is a state-action pair that achieves the minimum in $d_\text{min}$.
Thus, setting $B = d_\text{min}^{-1}$ can provide error bounds comparable to those in other works while ensuring global optimality.

Our goal is to establish the optimality of $d_e$ within the confidence set $D_B$.
The proof comprises two distinct steps.
Firstly, we establish that $w_{\tilde{d}}$ is feasible to the dual empirical LP~\eqref{eq:dual_offline} with high probability for any $\tilde{d} \in D_B$, under appropriate level of relaxation $\epsilon_x$.
Next, we show that $w_e$ has a (nearly) higher objective than $w_{\tilde{d}}$ with high probability since $w_e$ has a small duality gap.
In the following lemma, we prove that for any $\tilde{d} \in D_B$, corresponding $w_{\tilde{d}}$ is a feasible solution to the empirical LP with high probability under appropriate relaxation level $\epsilon_x$ in the constraint.
\begin{lemma}\label{lem:feasibility}
In dual empirical LP~\eqref{eq:dual_offline}, let 
\begin{equation}
\begin{split}
    \epsilon_x =  &\Bigg(B(1+\gamma)\gamma^H  + B(1+\gamma)(1-\gamma^H) \\
    &\times\sqrt{\frac{2|S||A|}{N}\log \frac{2N_x}{\delta}}\Bigg) \mathbf{1},    
\end{split}
\end{equation}
where $\delta>0$. Then, for any $\tilde{d} \in D_B$, $w_{\tilde{d}}$ is feasible to~\eqref{eq:dual_offline} with probability at least $1 - \frac{\delta}{2^{|S||A|}}$.
\end{lemma}
\begin{proof}
See Appendix~\ref{app:lem:feasibility}.
\end{proof}

From the above Lemma, we establish the optimality guarantee in the following theorem.
Specifically, in words, under the reward function $r$ recovered from the set \eqref{eq:solution}, we show that $d_e$ is an $\tilde{O}(\sqrt{|S||A|/N})$-suboptimal solution over the confidence set $D_B$ with high probability.

\begin{theorem}\label{thm:optimality}
Suppose $(u_\mathcal{D}, v_\mathcal{D}) \in \hat{\mathcal{R}}_\text{IRL}(\epsilon_g)$, with the relaxation level $\epsilon_x$ specified in Lemma~\ref{lem:feasibility}.
Let $r$ satisfy
\begin{equation}
r(s,a) = \frac{u_\mathcal{D}(s,a)}{\hat{d}_e(s,a)}
\end{equation}
for all $(s, a) \in S \times A$, following the convention $0/0=0$. Then, we have
\begin{equation}
   \mathbb{P} (r^\top d_e \geq r^\top \tilde{d} - \epsilon, \;\; \forall \tilde{d} \in D_B) \geq 1 - 3 \delta,
\end{equation}
where 
\begin{equation}
\begin{split}
    \epsilon = \epsilon_g + (1+B) \gamma^H + (1-\gamma^H)\sqrt{\frac{2}{N} \log \frac{1}{\delta}}\\
    +  B (1-\gamma^H)\sqrt{\frac{2|S||A|}{N} \log \frac{2}{\delta}}.    
\end{split}
\end{equation}
\end{theorem}
\begin{proof}
See Appendix~\ref{app:thm:optimality}.
\end{proof}

\paragraph{Sample complexity analysis.}
The proposed solution set achieves the $\tilde{O}(B(1-\gamma^H)\sqrt{|S||A|/N})$ sample complexity bound with additional error terms $\epsilon_g$ and $(1+B)\gamma^H$.
Note that the parameter $\epsilon_g$ can be set to $\epsilon_g = \tilde{O}(1/\sqrt{N})$ to match the sample complexity.
The term $(1+B)\gamma^H$ diminishes exponentially with the horizon of the collected trajectory data, which underscores the requirement for long-horizon data to ensure accurate estimation.
To the best of our knowledge, besides our work, \citep{zeng2023understanding} is the only other study that offers an optimality guarantee for the offline IRL problem under a discounted MDP, with a sample complexity of $\tilde{O}(1/\sqrt{N})$.
However, as their algorithm is based on a bi-level optimization approach and their error bound is given for the log-likelihood function, a direct comparison of their result with ours is not feasible.
The concurrent work by~\citep{zhao2023inverse} presents a comparable sample complexity for an episodic MDP. We provide a detailed comparison in Appendix~\ref{app:comparison}.

\paragraph{Trade-off between optimality and feasibility.}
In our formulation, there exists a trade-off between the optimality of the policy and the feasibility of the reward function, which is modulated by the parameter $\epsilon_g$.
The duality gap bound, denoted as $\epsilon_g$, adjusts the size of the solution set $\hat{\mathcal{R}}_\text{IRL}(\epsilon_g)$; this set expands with an increase in $\epsilon_g$.
$\epsilon_g$ can be reduced to $0$ without causing infeasibility, as the set $\hat{\mathcal{R}}_\text{IRL}(\epsilon_g)$ is always non-empty due to the trivial solution $(u, v) = (0, 0)$.
A smaller value of $\epsilon_g$ enhances the optimality of the expert policy $\pi_e$, as stated in Theorem~\ref{thm:optimality}.
However, excessively reducing $\epsilon_g$ can lead to overly greedy choices, resulting in trivial or degenerate solutions.
The impact of varying $\epsilon_g$ is demonstrated through numerical experiments in Section~\ref{sec:experiment}.

\paragraph{Function approximation.}
The proposed LP formulation is well-suited for function approximation (parameterization), which allows us to reduce both the computational cost and the size (dimension) of the solution set.
Consider the parameterization of the variable $(u, v)$ as $(u_\theta, v_\theta)$, where $\theta$ represents a parameter within the parameter space $\Theta \subset \mathbb{R}^k$, which we aim to explore.
If there exists a $\theta \in \Theta$ such that $(u_\theta, v_\theta) \in \hat{\mathcal{R}}_\text{IRL}(\epsilon_g)$, then the optimality guarantee provided in Theorem~\ref{thm:optimality} is preserved for the reward function recovered from $u_\theta$, while the computational complexity of the LP can be reduced to a polynomial in $k$, down from $|S||A|$.
It is important to note that this formulation remains a linear (or convex) program under linear (or convex) parameterization.
If non-convex function approximation is employed for high-dimensional or continuous state-action spaces, an efficient algorithm for solving the proposed optimization may be required; however, such an extension is beyond the scope of this work, and we defer this to future research.

\subsection{Degeneracy Issue in Reward Learning}\label{sec:IRLdegeneracy}
In the practical applications of reward learning, estimating the feasible reward set is not enough; we need to select a single reward function within the estimated feasible reward set to use.
This is not a trivial problem due to existence of degenerate reward functions in the feasible reward set.
Degenerate reward functions (e.g. $r=\mathbf{0}$), though theoretically feasible, are practically undesirable as they fail to separate the expert policy $\pi_e$ from others.
In our solution set $\hat{\mathcal{R}}_\text{IRL}(\epsilon_g)$~\eqref{eq:solution}, degeneracy in the feasibility constraint $K_\mathcal{D}^\top X v \geq u$ is critical.
If equality holds for some state-action pairs such that $(K_\mathcal{D}^\top X v - u)(s, a) = 0$, then the complementary slackness condition will not be violated by changing the value of $w_e(s,a)$, meaning that $w_e$ may not be uniquely optimal. 
We suggest a simple and tractable method to obtain a non-degenerate reward function in $\hat{\mathcal{R}}_\text{IRL}(\epsilon_g)$.

\paragraph{Utilizing suboptimal trajectory samples.}
A straightforward approach to obtain a non-degenerate solution is utilizing a suboptimal policy $\pi_\text{sub}$.
To be specific, we directly maximize the expected reward gap between the expert policy $\pi_e$ and the suboptimal $\pi_\text{sub}$.
A viable example of $\pi_\text{sub}$ is a uniformly random policy such as $\pi_\text{sub}(a|s) = \frac{1}{|A|}$ $\forall (s,a) \in S \times A$, because this policy is unlikely to be optimal unless the expected rewards are uniform over all actions.
To maximize the reward gap, we sample suboptimal trajectories with $\pi_\text{sub}$ and estimate the occupancy measure of $\pi_\text{sub}$ as $\hat{d}_\text{sub}$, using the sampling and estimation methods discussed previously. 
Then, we maximize the empirical mean of the reward gap as per the following LP:
\begin{equation}\label{eq:irl}
\begin{split}
\max_{r ,u, v} \quad &  r^\top  (\hat{d}_e -\hat{d}_\text{sub}) \\
\text{s.t.} \quad & (u, v) \in \hat{\mathcal{R}}_\text{IRL}(\epsilon_g), \quad u = \hat{d}_e \circ r.
\end{split}
\end{equation}
Here, $\circ$ denotes the element-wise (Hadamard) product.
The numerical experiments in Section~\ref{sec:experiment} demonstrate that the above formulation efficiently recovers a non-degenerate reward function with only a small number of suboptimal trajectory samples.


%% file: sections/section4.tex
\section{Offline Reinforcement Learning from Human Feedback}
\subsection{LP Formulation of Offline RLHF}\label{sec:RLHFformulation}
In this section, we extend our LP framework to address offline RLHF problem.
As discussed in Section~\ref{sec:formulation:feedback}, our focus is on minimizing the error associated with the reward $r$ and the human feedback data $\mathcal{D}_\text{HF}$.
We begin by representing the cumulative reward of each trajectory $\tau^{n, i}$ ($i=1, 2$) in the dataset $\mathcal{D}_\text{HF}$ as a linear function of the reward $r$.
Specifically, the cumulative reward from the trajectory $\tau^{n, i}$ can be expressed as $r(\tau^{n, i}) = r^\top \psi^{n, i}$, where each vector $\psi^{n, i} \in \mathbb{R}^{|S||A|}$ can be mapped from the trajectory $\tau^{n, i}$ by
\begin{equation}\label{eq:vectorization}
    \psi^{n,i} (s, a) :=  \sum_{h=0}^{H-1} \gamma^h \mathbf{1}_{\{s_h^{n, i} = s, a_h^{n, i} = a\}},
\end{equation}
for any $(s,a) \in S \times A$.
Following this, we define the error in the single data point $(\tau^{n, 1},  \tau^{n, 2}, y^n)$ associated with the reward function $r$ as
\begin{equation}
\begin{split}
\mathcal{L} (\tau^{n, 1},  \tau^{n, 2}, y^n ; r) :=  r^\top (\psi^{n, 2} - \psi^{n, 1}) \mathbf{1}_{\{y^n = 1\}}  \\
+ r^\top (\psi^{n, 1} - \psi^{n, 2}) \mathbf{1}_{\{y^n = 2\}}.    
\end{split}
\end{equation}

Note that naively minimizing the average or maximum error over queries might often lead to degenerate reward functions. 
This is because human evaluators sometimes provide conflicting feedback, such as $y^n=1$ when $r^\top \psi^{n, 2} > r^\top \psi^{n, 1}$, due to their stochasticity.
Under conflicting comparison data, minimizing $\mathcal{L}$ may result in degenerate solutions such as $r=\mathbf{0}$.
To address this issue, we allows for a slack in the error $\mathcal{L}$ by introducing a parameter $\epsilon_r \in \mathbb{R}$, which controls the size of the solution set.
Specifically, we define the solution set $\hat{\mathcal{R}}_\text{HF}$ as follows:
\begin{equation}\label{eq:RLHFset}
\begin{split}
    \hat{\mathcal{R}}_\text{HF}(\epsilon_r) :=  \{r \mid  &\mathcal{L} (\tau^{n, 1},  \tau^{n, 2}, y^n ; r)  \leq \epsilon_r\\
    &\forall n=1, 2, \ldots, N_q, \quad  r\in [-1, 1]^{|S||A|}\}.    
\end{split}
\end{equation}
Under this adjustable solution set, if we have additional information, we could also apply the strategy discussed in the previous section for identifying non-degenerate solutions: maximizing the reward gap between the expert trajectories and the suboptimal trajectories.

One advantage of the proposed LP method is its robustness to different human evaluator preference models, in contrast to MLE-based algorithms.
When the human evaluator deviates from the preference model assumed in MLE, the true reward parameter may diverge from the parameter space.
However, the LP approach is not subject to this limitation.
We illustrate this point with a bandit example in Appendix~\ref{app:mle}, where our LP algorithm successfully finds an optimal policy, whereas the recent pessimistic MLE algorithm fail to do so.

\subsection{Generalization Guarantee for Offline RLHF}
Recent works in offline RLHF, such as \citep{zhu2023principled, zhan2023provable}, have proposed a pessimistic MLE algorithm and provided an error bound between the estimated and the true reward function of the supposed preference model.
Our LP method does not offer an optimality guarantee in the same way as these works, as it obtains a set of reward functions without assuming a specific preference model.
Instead, we analyze the generalization property of the obtained reward functions by examining how $r \in \hat{\mathcal{R}}_\text{HF}(\epsilon_r)$ aligns with unseen trajectory pairs sampled from $\mu_\text{HF}$.

We first introduce the probabilistic preference model for a human evaluator generating feedback data.
We emphasize that our proposed method is not dependent on this model; we introduce it to analyze a generalization property.
Suppose that $y \in \{1, 2\}$ is sampled from a Bernoulli distribution with the probabilistic model $\mathbb{P}(y = 1 \mid \tau^{1}, \tau^{2}) = \Phi(r_\text{true}^\top (\psi^{1} - \psi^{2}))$, where $\Phi:\mathbb{R} \mapsto [0, 1]$ is a  monotonically non-decreasing function satisfying $\Phi(x) + \Phi(-x) = 1$ for all $x \in \mathbb{R}$. 
$\Phi$ represents the preference model of the evaluator, based on their personal reward function $r_\text{true}$.
For example, if $\Phi$ is a sigmoid function, i.e. $\Phi(x) = 1/(1+e^{-x})$, then the above probabilistic model is reduced to the Bradley-Terry-Luce (BTL) model~\citep{christiano2017deep}.
In the following theorem, we provide a generalization guarantee of any reward functions $r$ contained in the estimated solution set $\hat{\mathcal{R}}_\text{HF}(\epsilon_r)$.
Specifically, for a random (unseen) trajectory pair $(\tau^1, \tau^2)$ sampled from the sampling distribution $\mu_\text{HF}$ and the human feedback $y$ sampled from the preference model $\Phi$, we prove that the error $\mathcal{L} ( \tau^{1},  \tau^{2}, y; r)$ is bounded by $\epsilon_r$ with high probability.

\begin{figure*}[ht]
\begin{center}
\includegraphics[width=.48\textwidth]{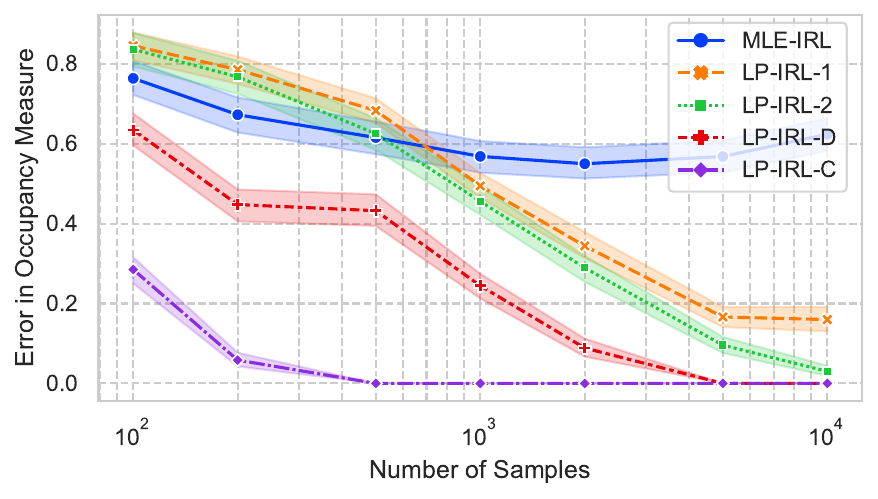}
\hfill
\includegraphics[width=.48\textwidth]{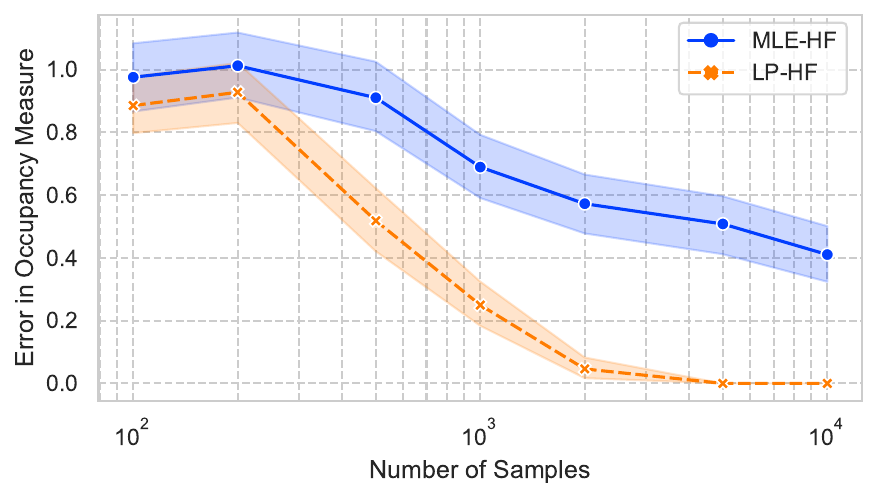}
\caption{$\mathcal{L}^1$ error in the optimal occupancy measure under an estimated reward function. Left: Offline IRL algorithms; Right: Offline RLHF algorithms.}
\label{fig:irl}
\end{center}
\vskip -0.2in
\end{figure*}

\begin{theorem}\label{thm:rlhf}
Suppose $r \in \hat{\mathcal{R}}_{\text{HF}}(\epsilon_r)$ and the human feedback data $(\tau^1, \tau^2, y) $ is sampled i.i.d. from the joint distribution $(\mu_{\text{HF}}, \Phi)$.
Then, for any $\delta \in (0, 1)$, 
\begin{equation}
     \mathbb{P} \left( \mathcal{L} ( \tau^{1},  \tau^{2}, y; r) \geq \epsilon_r \right) \leq \sqrt{\frac{1}{2 N_q} \log \frac{1}{\delta}}
\end{equation}
holds with probability at least $ 1-\delta $.
\end{theorem}
\begin{proof}
See Appendix~\ref{app:thm:rlhf}.
\end{proof}

Note that a trade-off between optimality and feasibility exists in $\hat{\mathcal{R}}_{\text{HF}}(\epsilon_r)$ with respect to the parameter $\epsilon_r$, similar to $\hat{\mathcal{R}}_{\text{IRL}}(\epsilon_g)$ in offline IRL.
Specifically, while it is preferable to set the parameter $\epsilon_r$ as small as possible to avoid overly relaxing the RLHF constraint, excessively reducing $\epsilon_r$ can lead to feasibility issues.
In practice, if any prior knowledge about the preference model $\Phi$ is available, this information can guide the selection of $\epsilon_r$.
If no such information is available, $\epsilon_r$ can be determined experimentally by starting with a large value and gradually reducing it until the reward set becomes trivial or infeasible.

\subsection{Integration of IRL and RLHF}\label{sec:integration}
Finally, our LP framework facilitates the integration of two types of expert data: IRL (trajectories collected from the expert policy) and RLHF (pairwise trajectory comparisons).
This is a unique feature of our LP framework, one that remains unexplored in the MLE framework.
We propose to recover the reward function $r$ from the intersection of two sets $\hat{\mathcal{R}}_\text{IRL}(\epsilon_g)$ and $\hat{\mathcal{R}}_\text{HF}(\epsilon_r)$ such that
\begin{equation}\label{eq:irl-rlhf:d}
\begin{split}
&\hat{\mathcal{R}}_\text{IRL-HF}(\epsilon_g, \epsilon_r) = \\
&\{ (r, u, v) \mid (u, v) \in \hat{\mathcal{R}}_\text{IRL}(\epsilon_g),\; r \in \hat{\mathcal{R}}_\text{HF}(\epsilon_r), \; u = \hat{d}_e \circ r\}.
\end{split}
\end{equation}
In this combined formulation, the IRL constraint $(u, v) \in \hat{\mathcal{R}}_\text{IRL}(\epsilon_g)$ provides the optimality guarantee of the expert policy, while the RLHF constraint $r \in \hat{\mathcal{R}}_\text{HF}(\epsilon_r)$ reduces the solution set and mitigates degeneracy by imposing additional constraints.

\paragraph{Strict trajectory comparison.}
If we can impose a strict reward gap between two trajectories, then the RLHF constraint can mitigate degeneracy more effectively.
For instance, the constraint $r(\tau^1) \geq r(\tau^2) + \delta$ eliminates degenerate solutions that satisfy $r(\tau^1) = r(\tau^2)$ from the solution set, if we can set a strict reward gap $\delta > 0$ based on human feedback data.
To formalize this, we extend our approach to include the continuous feedback case, wherein the feedback is given as a continuous variable, $y \in [-1, 1]$, instead of the discrete variable used in previous sections.
In Appendix~\ref{app:continuous}, we specify this continuous feedback setting, and suggest how we can extend the definition of the solution set $\hat{\mathcal{R}}_\text{HF}(\epsilon_r)$ as well as the generalization guarantee provided in Theorem~\ref{thm:rlhf}.
Additionally, in the next section, we compare the effects of discrete and continuous human feedback through numerical experiments.

%% file: sections/experiment.tex
\section{Numerical Experiments}\label{sec:experiment}
In this section, we demonstrate the performance of our LP algorithms through numerical experiments, comparing them to MLE algorithms in the literature.
We consider an MDP with $|S|=10$, $|A|=2$, and $\gamma=0.95$. In each experimental run, $P$ and $\mu_0$ are randomly selected.
To introduce additional complexity to the problem, we have set the true rewards to have similar values: $r_{\text{true}}(s, a_1) = 1.0$ and $r_{\text{true}}(s, a_2) = 0.9$ for all states $s\in S$.
The performance of each algorithm is assessed by measuring the proximity of an optimal occupancy measure under the true reward $r_\text{true}$ and the estimated reward function $\hat{r}$.
Specifically, we report $\Vert d^*(r_\text{true}) - d^*(\hat{r})\Vert_1$, which represents the $\mathcal{L}^1$ error between the optimal occupancy measures under $r_\text{true}$ and $\hat{r}$.
In each experiment, we sample $N$ trajectories with a horizon of $H=20$ according to $\pi_e$ in IRL, and $\mu_\text{HF}$ in RLHF.
For each sample size $N$, we conducted 200 experiments and reported the mean and standard deviation of the error.
See Appendix~\ref{app:experiment} for detailed parameters and algorithms used in the experiments.

\paragraph{Offline IRL.}
The left side of Figure~\ref{fig:irl} compares the errors associated with each IRL algorithm. The results indicate that our LP-based algorithms generally outperform the bi-level optimization-based MLE algorithm~\citep{zeng2023understanding}, demonstrating that LP is more sample-efficient in addressing ambiguity in the dynamics and the expert policy.
The solution set with a smaller relaxation level $\epsilon_g = 0.001/\sqrt{N}$ (LP-IRL-2) exhibits better performance than that with a greater relaxation level $\epsilon_g = 0.01/\sqrt{N}$ (LP-IRL-1).
This is consistent with the optimality-feasibility trade-off discussed in Section~\ref{sec:IRLoptimality}.
Additionally, the integration of IRL and RLHF data leads to improved performance, as predicted. The use of continuous feedback (LP-IRL-C) is even more effective than discrete feedback (LP-IRL-D) by facilitating stricter constraints.

\paragraph{Offline RLHF.}
In numerical experiments for offline RLHF, the human feedback data is generated following the greedy model.
The right side of Figure~\ref{fig:irl} compares the reward function obtained from LP~\eqref{eq:RLHFset} and the pessimistic MLE algorithm proposed by~\citep{zhu2023principled} under the BTL model.
In the LP algorithm, the error decreases rapidly as the number of samples increases, whereas the error in the MLE algorithm decreases more slowly.
This result is consistent with the discussion in Appendix~\ref{app:mle}, suggesting that the MLE algorithm might be inefficient or even fail if the human evaluator deviates from the assumed model, whereas LP does not.

%% file: sections/appendix.tex
\section{Related Work}~\label{app:relatedwork}
\vspace{-1em}
\paragraph{LP and Duality Approach in IRL.}
One of the foundational works in IRL \citep{ng2000algorithms} introduced the concept of characterizing a set of reward functions for which a given policy is optimal using an LP formulation.
This idea has been further developed in subsequent literature \citep{metelli2021provably, lindner2022active, metelli2023towards, zhao2023inverse}, as outlined in the introduction.
Recently proposed practical offline imitation learning (IL) algorithms, including ValueDICE \citep{kostrikov2019imitation}, IQ-Learn \citep{garg2021iq}, OPIRL \citep{hoshino2022opirl}, and ReCOIL \citep{sikchi2023dual}, address an occupancy matching problem that minimizes the statistical divergence between the learner and the expert distribution.
These algorithms exploit the duality of the LP formulation to obtain tractable algorithms, as extensively discussed in \citep{sikchi2023dual}.
Despite the practical success of these algorithms, the resulting reward function and policy depend on the model in use, and they lack theoretical performance guarantees, such as provable sample efficiency.

\paragraph{RLHF without Preference Model Assumption.}
In offline RLHF, we impose a margin-based constraint on the solution set, which allows for the alignment of reward functions with preference data without assuming any preference models of human evaluators. The concept of employing a margin constraint originated in the early imitation learning literature.
Specifically, maximum margin planning (MMP) \citep{ratliff2006maximum, ratliff2009learning} estimates the reward function such that the expert policy achieves a higher expected reward than all other policies by imposing a margin constraint in the reward optimization problem.
Recently, \citep{sikchi2023ranking} introduced Rank-Game, a two-player game formulation between a policy agent, which optimizes the policy given a reward function, and a reward agent, which aligns the reward function with offline pairwise preference data. Their algorithm is model-free, as the reward agent minimizes ranking loss without relying on a specific preference model. A unification of demonstration and preference data is also proposed in their work, similar to our approach in the LP framework.

\section{Proof of Lemma~\ref{lem:feasibility}}\label{app:lem:feasibility}
For ease of notation, let $\delta' = \frac{\delta}{2^{|S||A|}}$.
To show
\begin{equation}
\mathbb{P} (X^\top (K_\mathcal{D} w_{\tilde{d}} - (1-\gamma)\mu_0) \leq \epsilon_x) \geq 1-\delta',
\end{equation}
we divide our proof into two parts.
First, for any column $x_i$ of $X$, we show that
\begin{equation}
    x_i^\top (K_H - K) w_{\tilde{d}} \leq (1+\gamma) \gamma^H B
\end{equation}
holds for the matrix $K_H$, which will be defined later.
Next, we will prove that
\begin{equation}
     x_i^\top (K_\mathcal{D} - K_H) w_{\tilde{d}} \geq \epsilon_{xi} - (1+\gamma) \gamma^H B
\end{equation} holds with probability less than $\delta'/N_x$.
Then, combining both inequalities yields $x_i^\top (K_\mathcal{D} - K) w_{\tilde{d}} = x_i^\top (K_\mathcal{D} w_{\tilde{d}} - (1-\gamma)\mu_0) \geq \epsilon_{xi} $ holds with probability less than $\delta'/N_x$, since $K w_{\tilde{d}} = M \tilde{d} = (1-\gamma) \mu_0$ holds by $\tilde{d} \in D_B$.
Applying union bound to all columns $x_i$ of $X$ will lead to the conclusion.

To prove the first part, we first introduce the vector $d_e^H \in \mathbb{R}^{|S||A|}$, representing a finite-horizon truncation of $d_e$ up to the horizon $H-1$:
\begin{equation}
    d_e^H (s, a) := (1-\gamma) \sum_{h=0}^{H-1} \gamma^h P^{\pi_e}_{\mu_0} (s_h = s, a_h = a) \quad \forall (s,a) \in S \times A.
\end{equation}
We also define the vector $d_e^{P, H} \in \mathbb{R}^{|S|\times|A|\times|S|}$ as the truncation of $d_e'$ as follows:
\begin{equation}
d_e^{P, H} (s, a, s') := (1-\gamma) \sum_{h=0}^{H-1} \gamma^h P^{\pi_e}_{\mu_0} (s_h = s, a_h = a, s_{h+1} = s') \quad \forall (s,a, s') \in S \times A \times S.
\end{equation}
Then, we define the matrix $K_H \in \mathbb{R}^{|S|^2|A|}$ using $d_{e}^H$ and $d_e^{P, H}$ as follows:
\begin{equation}
K_H(s', (s,a)) := d_e^H(s,a) \mathbf{1}_{\{s= s'\}} - \gamma d_e^{P, H} (s, a, s')  \quad \forall (s,a, s') \in S \times A \times S.
\end{equation}
Since $K_H$ can be considered as a finite-horizon truncation of $K$ by its definition, only the terms from $h=H$ to $\infty$ remain in the matrix $K-K_H$.
Consequently, we get the following inequalities that prove the first part:
\begin{equation}
\begin{split}
    |x_i^\top (K - K_H) w_{\tilde{d}}| &\leq \sum_{s'\in S} |x_i(s')| \sum_{h=H}^\infty (1-\gamma) \gamma^h \sum_{a \in A} \mathbb{P}^{\pi_e}_{\mu_0} (s_h = s', a_h = a) w_{\tilde{d}} (s',a)  \\
    &+ \sum_{s'\in S} |x_i(s')| \sum_{h=H}^\infty (1-\gamma) \gamma^{h+1}  \sum_{(s,a) \in S \times A} \mathbb{P}^{\pi_e}_{\mu_0} (s_h = s, a_h = a, s_{h+1} = s') w_{\tilde{d}} (s,a) \\
    &\leq \sum_{s'\in S}\sum_{h=H}^\infty (1-\gamma) \gamma^h \mathbb{P}^{\pi_e}_{\mu_0} (s_h = s') B + \sum_{s'\in S}\sum_{h=H}^\infty (1-\gamma) \gamma^{h+1} \mathbb{P}^{\pi_e}_{\mu_0} (s_{h+1} = s') B \\
    &= \gamma^H B + \gamma^{H+1}B = (1+\gamma) \gamma^H B.
\end{split}
\end{equation}
The first inequality holds directly from the definitions of $K$ and $K_H$, and the second inequality results from the assumptions $\Vert x_i \Vert_\infty \leq 1$ and $\Vert w_{\tilde{d}} \Vert_\infty \leq B$.
Then, we have $x_i^\top (K_H - K) w_{\tilde{d}} \leq (1+\gamma) \gamma^H B$.

For the second step, consider the following definition of the random variable $z (\tau) \in \mathbb{R}$, where $\tau$ represents the finite-horizon trajectory sample:
\begin{equation}
\begin{split}
    z(\tau) :=  &\sum_{s' \in S} x_i(s') \sum_{(s,a) \in S \times A} w_{\tilde{d}} (s,a)(1-\gamma) \times \sum_{h=0}^{H-1} \left[  \gamma^h  \mathbf{1}_{\{s_h = s, a_h = a\}} - \gamma^{h+1} \mathbf{1}_{\{s_h = s, a_h = a, s_{h+1} = s'\}} \right].
\end{split}
\end{equation}
Then, $x_i^\top K_\mathcal{D} w_{\tilde{d}} = \frac{1}{N} \sum_{n=1}^N z(\tau^n)$ holds, implying that $x_i^\top K_\mathcal{D} w_{\tilde{d}}$ is the empirical mean of the random variable $z(\tau)$, derived from $N$ trajectory samples $\tau^n$.
Meanwhile, $x_i^\top K_H w_{\tilde{d}}$ represents the expected value of $z(\tau)$ over $\tau$.
Moreover, we can show that the random variable $z(\tau)$ has a bounded range as follows:
\begin{equation}
\begin{split}
    |z(\tau)| &\leq  B (1-\gamma) \sum_{s'\in S} \sum_{(s,a) \in S \times A}\sum_{h=0}^{H-1} \left[ \gamma^h \mathbf{1}_{\{s_h = s, a_h = a\}} + \gamma^{h+1} \mathbf{1}_{\{s_h = s, a_h = a, s_{h+1} = s'\}} \right] \\
    &\leq B (1-\gamma) \left( \frac{1-\gamma^H}{1-\gamma} + \frac{\gamma-\gamma^{H+1}}{1-\gamma}\right) = B(1+\gamma)(1-\gamma^H),
\end{split}
\end{equation}
where we used the assumptions $\Vert x_i \Vert_\infty \leq 1$ and $\Vert w_{\tilde{d}} \Vert_\infty \leq B$ in the first inequality.
Therefore, we can apply Hoeffding's inequality as follows:
\begin{equation}
    \mathbb{P} \left(x_i^\top (K_\mathcal{D} - K_H) w_{\tilde{d}} \geq \epsilon \right) \leq \exp \left( - \frac{N \epsilon^2}{2B^2 (1 + \gamma)^2 (1-\gamma^H)^2} \right) \quad \forall \epsilon \geq 0.
\end{equation}
Let $\epsilon=  \sqrt{\frac{2B^2 (1+\gamma)^2(1-\gamma^H)^2}{N}\log \frac{N_x}{\delta'}}$ and $\epsilon_{xi} = \epsilon + (1+\gamma) \gamma^H B$.
Then, the above inequality is equivalent to
\begin{equation}
    \mathbb{P} \left(x_i^\top (K_\mathcal{D} - K_H) w_{\tilde{d}} \geq \epsilon_{xi} - (1+\gamma) \gamma^H B\right) \leq \frac{\delta'}{N_x}.
\end{equation}
Plugging the inequality $x_i^\top (K_H - K) w_{\tilde{d}} \leq (1+\gamma) \gamma^H B$ derived in the first part, we get
\begin{equation}
    \mathbb{P} \left(x_i^\top (K_\mathcal{D} - K_H) w_{\tilde{d}} + x_i^\top (K_H - K) w_{\tilde{d}}  \geq \epsilon_{xi}  \right) \leq \mathbb{P} \left(x_i^\top (K_\mathcal{D} - K_H) w_{\tilde{d}} + (1+\gamma) \gamma^H B  \geq \epsilon_{xi}\right) \leq \frac{\delta'}{N_x}.
\end{equation}
Thus, $\mathbb{P} \left(x_i^\top (K_\mathcal{D} - K) w_{\tilde{d}} \geq \epsilon_{xi}\right) \leq \delta'/N_x$.
Taking the union bound to all events over $i=1, 2, \ldots, N_x$,
\begin{equation}
    \mathbb{P} \left(X^\top (K_\mathcal{D} - K) w_{\tilde{d}} \geq \epsilon_x\right) \leq \delta'.
\end{equation}
Since $Kw_{\tilde{d}} =  M \tilde{d} = (1-\gamma)\mu_0$ by $\tilde{d} \in D_B$, the above inequality is equivalent to 
\begin{equation}
    \mathbb{P} \left(X^\top (K_\mathcal{D}w_{\tilde{d}} - (1-\gamma)\mu_0) \geq \epsilon_x\right) \leq \delta',
\end{equation}
which completes the proof.

\section{Proof of Theorem~\ref{thm:optimality}}\label{app:thm:optimality}
The proof is comprised of three steps.
In the first step, we employ a concentration bound to establish a limit on the difference between $u^\top \mathbf{1}$ and $u_\mathcal{D}^\top \mathbf{1}$, ensuring that $\mathbb{P} (u^\top \mathbf{1} - u_\mathcal{D}^\top \mathbf{1} \geq - \epsilon_{u1}) \geq 1-\delta$ for a certain $\epsilon_{u1}$.
Next, from the feasibility of $w_{\tilde{d}}$ proven by Lemma~\ref{lem:feasibility} and using the optimality conditions in~\eqref{eq:solution}, we can show that $\mathbb{P}( u_\mathcal{D}^\top \mathbf{1} \geq u_\mathcal{D}^\top w_{\tilde{d}} - \epsilon_g, \; \forall  \tilde{d} \in D_B) \geq 1-\delta$ holds.
The final step is to bound the difference between $u_\mathcal{D}^\top w_{\tilde{d}}$ and $u^\top w_{\tilde{d}}$, showing that $\mathbb{P}(u_\mathcal{D}^\top w_{\tilde{d}} - u^\top w_{\tilde{d}} \geq  - \epsilon_{u2}, \; \forall \tilde{d} \in D_B) \geq 1 - \delta$ for a specific $\epsilon_{u2}$.
Combining these three results with the  union bound completes the proof.

We first prove that $\mathbb{P} (u^\top \mathbf{1} - u_\mathcal{D}^\top\mathbf{1}  \geq  - \epsilon_{u1}) \geq 1-\delta$ holds if we let $\epsilon_{u1}  = \sqrt{\frac{2(1-\gamma^H)^2}{N} \log \frac{1}{\delta}} + \gamma^H$.
Since $u^\top \mathbf{1} - u_\mathcal{D}^\top \mathbf{1} = r^\top d_e - r^\top \hat{d}_e = r^\top(d_e -d_e^H) + r^\top (d_e^H - \hat{d}_e)$,
we bound two terms $r^\top(d_e -d_e^H)$ and $r^\top (d_e^H - \hat{d}_e)$ separately.
First, $r^\top(d_e -d_e^H)$ can be bounded as
\begin{equation}
\begin{split}
|r^\top(d_e -d_e^H)| &\leq (1-\gamma) \sum_{(s,a) \in S \times A} |r(s,a)| \sum_{h=H}^{\infty} \gamma^h P^\pi_{\mu_0} (s_h = s, a_h = a)\\
&\leq (1-\gamma) \sum_{(s,a) \in S \times A} \sum_{h=H}^{\infty} \gamma^h P^\pi_{\mu_0} (s_h = s, a_h = a) = \gamma^H.
\end{split}
\end{equation}
Next, consider the random variable $z'(\tau)$ defined as
\begin{equation}
z'(\tau) := (1-\gamma) \sum_{(s,a) \in S \times A} r(s,a) \sum_{h=0}^{H-1} \gamma^h \mathbf{1}_{\{s_h = s, a_h = a\}},
\end{equation}
which represents the cumulative reward of $\tau$ multiplied by the constant $(1-\gamma)$.
According to its definition, $r^\top d_e^H$ is the expected value of $z'(\tau)$ over $\tau$, while $r^\top \hat{d}_e$ is the empirical mean of $z'(\tau)$ derived from the samples $(\tau^1, \tau^2, \ldots, \tau^N)$.
Moreover, from its definition, we can easily show that $z'(\tau)$ lies in the interval $[-(1-\gamma^H) , 1-\gamma^H]$.
Thus, we can apply Hoeffding's inequality to bound the term $r^\top (d_e^H - \hat{d}_e)$ as follows:
\begin{equation}
\mathbb{P} (r^\top (d_e^H - \hat{d}_e) \leq - \epsilon) \leq \exp \left(\frac{-N \epsilon^2}{2 (1-\gamma^H)^2} \right) \quad \forall \epsilon \geq 0.
\end{equation}
Letting  $\epsilon = \epsilon_{u1} - \gamma^H$ yields $\mathbb{P} (r^\top (d_e^H - \hat{d}_e) \leq -\epsilon_{u1} + \gamma^H) \leq  \delta$.
Then, adding two results completes the first steps follows:
\begin{equation}
\begin{split}
\mathbb{P} (u^\top \mathbf{1} - u_\mathcal{D}^\top \mathbf{1} \leq -\epsilon_{u1}) &= \mathbb{P} (r^\top (d_e^H - \hat{d}_e) + r^\top (d_e - d_H) \leq -\epsilon_{u1})\\
&\leq \mathbb{P} (r^\top (d_e^H - \hat{d}_e) \leq -\epsilon_{u1} + \gamma^H) \leq  \delta.
\end{split}
\end{equation}

Next, by Lemma~\ref{lem:feasibility}, $w_{\tilde{d}}$ is a feasible solution to \eqref{eq:dual_offline} with probability at least $1-\frac{\delta}{2^{|S||A|}}$.
If $w_{\tilde{d}}$ is a feasible solution to \eqref{eq:dual_offline}, then the following inequalities hold:
\begin{equation}
\begin{split}
    u_\mathcal{D}^\top \mathbf{1} & \overset{(i)}{\geq} (1-\gamma) \mu_0^\top X v_\mathcal{D}  + \epsilon_x^\top v_\mathcal{D} - \epsilon_g\\
    &\overset{(ii)}{\geq} w_{\tilde{d}}^{\top} K_\mathcal{D}^\top X v_\mathcal{D} - \epsilon_g\\
    &\overset{(iii)}{\geq} w_{\tilde{d}}^{\top} u_\mathcal{D} - \epsilon_g\\
\end{split}
\end{equation}
The inequality $(i)$ holds by the duality gap constraint, $(ii)$ holds because $w_{\tilde{d}}$ is feasible to \eqref{eq:dual_offline}, and $(iii)$ holds by the feasibility constraint $K_\mathcal{D}^\top X v_\mathcal{D} \geq u_\mathcal{D}$ and $w_{\tilde{d}}\geq 0$.
Therefore, we get
\begin{equation}
\mathbb{P} ( u_\mathcal{D}^\top \mathbf{1} \geq u_\mathcal{D}^\top w_{\tilde{d}} - \epsilon_g  ) \geq 1 - \frac{\delta}{2^{|S||A|}}.
\end{equation}
We then take union bound over all extreme points of $D_B$.
Since $D_B$ has at most $2^{|S||A|}$ extreme points, we get
\begin{equation}
\mathbb{P} ( u_\mathcal{D}^\top \mathbf{1} \geq u_\mathcal{D}^\top w_{\tilde{d}} - \epsilon_g, \; \forall \tilde{d} \in D_B  ) \geq 1 - \delta.
\end{equation}

In addition, by similar steps to the first part of the proof, we can show that
\begin{equation}
\mathbb{P} ( u_\mathcal{D}^\top w_{\tilde{d}} - u^\top w_{\tilde{d}} \geq  - \epsilon_{u2}  ) \geq 1 - \frac{\delta}{2^{|S||A|}},
\end{equation}
if we let $\epsilon_{u2} = B\sqrt{\frac{2(1-\gamma^H)^2|S||A|}{N} \log \frac{2}{\delta}} + B\gamma^H$.
Taking union bound over all extreme points of $\tilde{d} \in D_B$ yields
\begin{equation}
\mathbb{P} ( u_\mathcal{D}^\top w_{\tilde{d}} - u^\top w_{\tilde{d}} \geq  - \epsilon_{u2}, \; \forall \tilde{d} \in D_B  ) \geq 1 - \delta,
\end{equation}
Taking union bound to the above three cases and using $u^\top \mathbf{1} = r^\top d_e$ and $u^\top w_{\tilde{d}}  = r^\top \tilde{d}$ from Lemma~\ref{lem:substitution}, the conclusion holds as
\begin{equation}
    \mathbb{P} (r^\top d_e \geq r^\top \tilde{d} - \epsilon, \; \forall \tilde{d} \in D_B )
 = \mathbb{P} (u^\top \mathbf{1}\geq u^\top w_{\tilde{d}}  - \epsilon_g - \epsilon_{u1} - \epsilon_{u2}, \; \forall \tilde{d} \in D_B ) \geq 1 - 3\delta.
\end{equation}

\section{Comparison to Pessimism-based Approach}\label{app:comparison}
The concurrent work by~\citep{zhao2023inverse} proposed an offline IRL algorithm for finite-horizon MDPs with comparable sample complexity, based on pessimistic value iteration.
To be specific, they recover the mapping from the value and advantage functions to reward functions through Bellman iterations under estimated state transition probabilities.
Though a direct comparison is limited since our work is developed for infinite-horizon discounted MDPs, there are some common structures between their algorithm and ours.
Specifically, the value and advantage functions in their work can be considered as the primal optimization variable and the slack in the primal feasibility constraint in our formulation.

Nevertheless, there are differences between the resulting reward functions from their algorithm and ours.
To address the uncertainty caused by non-uniform data coverage in the offline setting, they penalize the reward on uncertain state-action pairs that are less visited in the dataset. Such a pessimism-based reward estimation framework provides strong theoretical optimality guarantees, such as finite sample complexity bounds, similar to our approach. However, in contrast to our solution set, which is a polyhedron, the use of a nonlinear and non-convex penalty function in their reward model leads to a solution set that is also nonlinear and non-convex. This distinction makes our algorithm more flexible for any extension, such as function approximation and the integration of additional information.

\section{Proof of Theorem~\ref{thm:rlhf}}\label{app:thm:rlhf}
We first define  the random variable 
\begin{equation}
    g(\tau^1, \tau^2, y) := \mathbf{1}_{\{\mathcal{L}(\tau^1, \tau^2, y; r) \geq \epsilon_r\}}(\tau^1, \tau^2, y),
\end{equation}
where $\mathbf{1}_{\{\mathcal{L}(\tau^1, \tau^2, y; r) \geq \epsilon_r\}}(\tau^1, \tau^2, y)$ is the  indicator function for the event that an error exceeds $\epsilon_r$, i.e. $\mathcal{L}(\tau^1, \tau^2, y; r) \geq \epsilon_r$.
The expected value of $g$ can be expressed as
\begin{equation}
\bar{g} = \mathbb{E}_{(\tau^1, \tau^2, y)\sim (\mu_{\text{HF}}, \Phi)} [g(\tau^1, \tau^2, y)] = \mathbb{P}(\mathcal{L}(\tau^1, \tau^2, y; r) \geq \epsilon_r).
\end{equation}
From the assumption that $ r \in  \hat{\mathcal{R}}_{\text{HF}}(\epsilon_r) $, the empirical mean of $g$ is given by $0$:
\begin{equation}
\hat{g} = \frac{1}{N_q} \sum_{n=1}^{N_q} g(\tau^{n, 1}, \tau^{n, 2}, y^n) = 0.
\end{equation}
From Hoeffding's inequality, we have $ \mathbb{P}(\hat{g} - \bar{g} \leq  -\epsilon) \leq e^{-2N_q \epsilon^2} = \delta $ if $ \epsilon =  \sqrt{\frac{1}{2 N_q} \log \frac{1}{\delta}}$.
Therefore, $\bar{g} \leq \epsilon$ holds with probability at least $1-\delta$, which completes the proof.

\section{Robustness of LP Framework}\label{app:mle}
In this section, we discuss the robustness of the proposed LP framework compared to MLE framework, with respect to the different preference models of human evaluators.
We first introduce the MLE framework in offline RLHF.
Consider the reward parameterization $r_\theta$, where $\theta \in \Theta$ is a parameter and  $\Theta \subset \mathbb{R}^k$ is a parameter space we aim to search the optimal parameter.
Then, the standard MLE framework can be illustrated as maximizing the log-likelihood function as follows:
\begin{equation}
    \hat{\theta}_{\text{MLE}} \in \argmax_{\theta \in \Theta} \sum_{n=1}^{N_q} \log \left( \Phi \left(- \mathcal{L} ( \tau^{n, 1},  \tau^{n, 2}, y^n; r_\theta)\right)\right).
\end{equation}
Our LP framework finds the reward parameter in the solution set, such that $\hat{\theta}_{\text{LP}} \in \{\theta \in \Theta \mid r_\theta \in \hat{\mathcal{R}}_\text{HF}(\epsilon_r) \}$.
Under the estimated reward parameters, we obtain the policy maximizing the reward function as follows:
\begin{equation}
    \hat{\pi}_{\text{LP}} \in \argmax_\pi \mathbb{E}_{s\sim d^\pi} [r_{\hat{\theta}_\text{LP}}(s, \pi(s))], \quad \hat{\pi}_{\text{MLE}} \in \argmax_\pi \mathbb{E}_{s\sim d^\pi} [r_{\hat{\theta}_\text{MLE}}(s, \pi(s))].
\end{equation}

\paragraph{Pessimistic MLE.}
Recent works in offline RLHF, such as those by~\citep{zhan2023provable, zhu2023principled} have adapted the concept of pessimism from offline RL theory to address the data coverage issue.
Specifically, these studies define a confidence set for the reward function and solve robust optimization problem to identify the policy that maximizes the worst-case reward within this set.
For example, \citep{zhu2023principled} uses a semi-norm $\Vert \cdot \Vert_{\Sigma + \lambda I}$ as a metric for constructing the confidence set, where $\Sigma$ represents the covariance of the comparison data and $\lambda>0$ is a conservativeness parameter, such that
\begin{equation}
    \mathcal{D}_\text{PE} =  \left\{\theta \in \Theta  \mid  \Vert \theta - \hat{\theta}_\text{MLE}\Vert_{\Sigma + \lambda I} \leq f(N, k, \delta, \lambda) \right\}.
\end{equation}
Then, the policy is optimized for the worst-case parameter in $ \mathcal{D}_\text{PE}$, such that
\begin{equation}
    \hat{\pi}_{\text{PE}} \in \argmax_\pi \min_{\theta \in \mathcal{D}_\text{PE}}  \mathbb{E}_{s\sim d^\pi} [r_\theta(s, \pi(s))].
\end{equation}
\citep{zhu2023principled} prove that the true parameter $\theta^*$ exists in this confidence set with high probability as the number of sample increases, which enables them to provide an optimality guarantee.

However, the MLE-based algorithms require an assumption that a human evaluator follows a specific preference model, and the true reward parameter corresponding to the model should lie in the parameter space, such that $\theta^* \in \Theta$, where $\Theta$ must be bounded.
Such realizability assumption can easily be violated in practice, when the true preference model deviates from the model used in algorithm.
For instance, if the BTL model is assumed but the human evaluator follows the greedy policy, the true parameters diverge to $+\infty$ or $-\infty$, which violates the assumption that $\theta^* \in \Theta$.
To illustrate these points, we provide a simple bandit problem that both MLE and pessimistic MLE fail but LP succeeds to recover the optimal policy.

\begin{proposition}\label{prop:mle}
For any $\delta >0$, there exists a linear bandit and a sampling distribution $\mu_\text{HF}$ such that
\[
\hat{\pi}_\text{LP} = \pi^*, \quad \hat{\pi}_\text{MLE} \neq \pi^*, \quad \text{and} \quad \hat{\pi}_\text{PE} \neq \pi^*
\]
hold with probability at least $1-\delta$.
\end{proposition}
\begin{proof}
Consider a linear bandit with a single state $s$ and three actions $a_1$, $a_2$, and $a_3$.
We consider the tabular setting such that $r_\theta = [\theta_1, \theta_2, \theta_3]$, where $\theta_i$ denotes the reward for the action $a_i$.
Suppose that human evaluators follow the deterministic (greedy) model, and the preference order is given by $a_3 > a_2 > a_1$, i.e. $a_3$ is the most preferable and $a_1$ is the least preferable action.

We construct a sampling distribution such that both MLE and pessimistic MLE algorithms in \citep{zhu2023principled} returns a wrong policy with high probability, while LP succeeds to find an optimal policy.
Specifically, if the pair $(a_1, a_2)$ is sampled with a significantly higher probability compared to the pair $(a_2, a_3)$ in the queries, we show that $\{\hat{\theta}_{\text{MLE}}\}_2 > \{\hat{\theta}_{\text{MLE}}\}_3$ holds under the BTL model and the greedy evaluator.
The MLE algorithm proposed in~\citep{zhu2023principled} estimate the reward parameter by solving 
\begin{equation}
\hat{\theta}_{\text{MLE}} \in \argmax_{\theta \in \Theta} \frac{N_{12}}{N} \log \frac{e^{\theta_2}}{e^{\theta_1} + e^{\theta_2}}  + \frac{N_{23}}{N} \log \frac{e^{\theta_3}}{e^{\theta_2} + e^{\theta_3}} + \frac{N_{31}}{N} \log \frac{e^{\theta_3}}{e^{\theta_3} + e^{\theta_1}},
\end{equation}
where $N_{ij}$ denotes the number of queries $(a_i, a_j)$, $N = N_{12} + N_{23} + N_{31}$, and $\Theta = \{\theta \mid \mathbf{1}^\top \theta = 0, \Vert \theta \Vert_2 \leq 1\}$.
We first prove the following lemma:
\begin{lemma}\label{lem:mle}
Let $\theta^* = [\theta^*_1, \theta^*_2, \theta^*_3]$ be an optimal solution to the following optimization problem:
\begin{equation}
\begin{split}
    &\max_{\theta \in \Theta} \; J(\theta_1, \theta_2, \theta_3) = \alpha \log \frac{e^{\theta_2}}{e^{\theta_1} + e^{\theta_2}} + \beta \log \frac{e^{\theta_3}}{e^{\theta_2} + e^{\theta_3}} \\
\end{split}
\end{equation}
where $\alpha, \beta \in (0, 1)$.
If $\alpha > 2 e^3 \beta$, then $\theta^*_2 > \theta^*_3$.
\end{lemma}
\begin{proof}
Define the Lagrangian function $\mathcal{L}(\theta_1, \theta_2, \theta_3, \lambda_1, \lambda_2) = J(\theta_1, \theta_2, \theta_3) + \lambda_1 (1- \theta_1^2 - \theta_2^2 - \theta_3^2) + \lambda_2 (\theta_1 + \theta_2 + \theta_3)$.
From the KKT conditions,
\begin{equation}
\begin{split}
    \left[ \frac{\partial \mathcal{L}}{\partial \theta_2} \right]_{(\theta^*, \lambda^*)} &= \alpha \frac{e^{\theta^*_1}}{e^{\theta^*_1} + e^{\theta^*_2}} - \beta \frac{e^{\theta^*_2}}{e^{\theta^*_2} + e^{\theta^*_3}} - 2\lambda^*_1 \theta^*_2 + \lambda^*_2 = 0,\\
    \left[ \frac{\partial \mathcal{L}}{\partial \theta_3} \right]_{(\theta^*, \lambda^*)} &=  \beta \frac{e^{\theta^*_2}}{e^{\theta^*_2} + e^{\theta^*_3}} - 2\lambda^*_1 \theta^*_3 + \lambda^*_2 = 0.
\end{split}
\end{equation}
Subtracting both equations yields
\begin{equation}
    \alpha \frac{e^{\theta^*_1}}{e^{\theta^*_1} + e^{\theta^*_2}} - 2\beta \frac{e^{\theta^*_2}}{e^{\theta^*_2} + e^{\theta^*_3}}  = 2\lambda^*_1 (\theta^*_2 - \theta^*_3). 
\end{equation}
If $\alpha > 2 e^3 \beta$, we can show that the left hand side of the above equality must be greater than $0$ as follows:
\begin{equation}
\begin{split}
     \alpha \frac{e^{\theta^*_1}}{e^{\theta^*_1} + e^{\theta^*_2}} - 2\beta \frac{e^{\theta^*_2}}{e^{\theta^*_2} + e^{\theta^*_3}} &> 2 \beta \left(\frac{e^{\theta^*_1 + 3}}{e^{\theta^*_1} + e^{\theta^*_2}} - \frac{e^{\theta^*_2}}{e^{\theta^*_2} + e^{\theta^*_3}} \right) = 2 \beta \frac{e^{\theta^*_1 + \theta^*_2 + 3} + e^{\theta^*_1 + \theta^*_3 + 3} - e^{\theta^*_1 + \theta^*_2} - e^{2\theta^*_2}}{(e^{\theta^*_1} + e^{\theta^*_2})(e^{\theta^*_2} + e^{\theta^*_3})}\\
     &=2 \beta \frac{(e^{\theta^*_1 + \theta^*_2 + 3}- e^{\theta^*_1 + \theta^*_2})  + (e^{3-\theta^*_2}  - e^{2\theta^*_2})}{(e^{\theta^*_1} + e^{\theta^*_2})(e^{\theta^*_2} + e^{\theta^*_3})} > 0,
\end{split}
\end{equation}
where the last inequality comes from $\theta^*_2 \leq 1$.
Then, the right hand side $2\lambda^*_1 (\theta^*_2 - \theta^*_3)$ must be greater than zero as well.
Since $\lambda_1^*\geq 0$, we get $\theta^*_2 > \theta^*_3$.
\end{proof}

By Lemma~\ref{lem:mle}, there exist $\alpha, \beta \in (0, 1)$ such that if $ \frac{N_{12}}{N} \geq \alpha$, $\frac{N_{23}}{N} \leq \beta$, and $N_{31} = 0$, then $\{\hat{\theta}_{\text{MLE}}\}_2 > \{\hat{\theta}_{\text{MLE}}\}_3$.
For any $\delta>0$, there exists a sampling distribution $\mu_\text{HF}$ satisfying
\begin{equation}
 \mathbb{P} (N_{12} \geq \alpha N , \; 1\leq N_{23} \leq \beta N, \; N_{31} = 0 ) \geq 1- \delta.
\end{equation}
Then, under this sampling distribution $\mu_\text{HF}$, $\hat{\pi}_{\text{MLE}} (s) = a_2$ with probability at least $1-\delta$, while $\pi^*(s) = a_3$.

Next, we consider the pessimistic MLE under $\mu_\text{HF}$.
The pessimistic MLE imposes higher penalty on the reward function of state-action pairs that have less support in the data.
Therefore, intuitively, the penalty for the state $a_3$ will be higher than $a_2$, and thus, $\hat{\pi}_\text{PE}(s) = a_2$ will hold.
We use the penalty function proposed in \citep{zhu2023principled} to confirm this.
Consider the covariance matrix
\begin{equation}
    \Sigma = \frac{1}{N} \begin{bmatrix}
  N_{12}  & -N_{12} & 0\\
  -N_{12} & N_{12} + N_{23} & -N_{23} \\
  0 & -N_{23} & N_{23}
\end{bmatrix}.
\end{equation}
Then, the penalty function for $a_2$ and $a_3$ are computed as
\begin{equation}
\phi_2 = \Vert [0, 1, 0] \Vert_{(\Sigma + \lambda I)^{-1}}^2 =  \frac{(N_{12} + \lambda)(N_{23} + \lambda)}{|\Sigma + \lambda I|}, \quad \phi_3=  \Vert [0, 0, 1] \Vert_{(\Sigma + \lambda I)^{-1}}^2   = \frac{ (N_{12} + \lambda)(N_{12} + N_{23} + \lambda) - N_{12}^2}{|\Sigma + \lambda I|}.
\end{equation}
It is easy to show that $\phi_3 \geq \phi_2$ for any $\lambda \geq 0$.
Then, the inequality 
\begin{equation}
\{\hat{\theta}_\text{MLE}\}_2 - c\Vert\phi_2\Vert_{(\Sigma + \lambda I)^{-1}} > \{\hat{\theta}_\text{MLE}\}_3 - c \Vert\phi_3\Vert_{(\Sigma + \lambda I)^{-1}}   
\end{equation}
holds for any constant $c>0$, and thus, $\hat{\pi}_{\text{PE}}$ chooses $a_2$ as the best action. Therefore, $\hat{\pi}_\text{PE} \neq \pi^*$.
Finally, since there exists at least one query of $(a_2, a_3)$ (by $N_{23} \geq 1$), we have $\{\hat{\theta}_{\text{LP}}\}_2 \leq \{\hat{\theta}_{\text{LP}}\}_3 + \epsilon_r$.
Let $\epsilon_r \leq 0$, we have $\hat{\pi}_\text{LP} = \pi^*$, which completes the proof.
\end{proof}

\section{Extension to Continuous Feedback}\label{app:continuous}
Suppose that the cumulative distribution function (CDF) for $y$, given a query pair $(\tau^{1}, \tau^{2})$, can be expressed as:
\begin{equation}
\mathbb{P}( y \leq \alpha \mid (\tau^{1}, \tau^{2})) = \Phi(\alpha; r_\text{true}^\top (\psi^{1} - \psi^{2})) \quad \forall \alpha \in [-1, 1].
\end{equation}
Specifically, $\Phi(\cdot; r) :[-1, 1] \mapsto [0, 1]$ is assumed to be a CDF for any $r \in \mathbb{R}$, i.e. right-continuous, monotonically non-decreasing, $\Phi(-1; r)=0$, and $\Phi(1; r)=1$. Furthermore, we assume that $\Phi(\alpha; \cdot):\mathbb{R} \mapsto [0, 1]$ is monotonically non-increasing with respect to $r$ to reflect a preference in the pairwise comparison.
Then, we enforce the reward gap to be greater than $c|y|$ by defining an error as
\begin{equation}
\mathcal{L}' ( \tau^{n, 1},  \tau^{n, 2}, y^n; r) :=  \left( cy^n + r^\top (\psi^{n, 2} - \psi^{n, 1}) \right) \mathbf{1}_{\{y^n \geq 0\}} + \left(-cy^n + r^\top (\psi^{n, 1} - \psi^{n, 2}) \right) \mathbf{1}_{\{y^n \leq 0\}},    
\end{equation}
where $c>0$ is a scaling parameter.
The solution set $\hat{\mathcal{R}}_\text{CHF}(\epsilon_r)$ is then defined in the same way with \eqref{eq:RLHFset} using the error $\mathcal{L}'$.
\begin{equation}
    \hat{\mathcal{R}}_\text{CHF}(\epsilon_r) :=  \{r \mid  \mathcal{L}' ( \tau^{n, 1},  \tau^{n, 2}, y^n; r)  \leq \epsilon_r \quad \forall n=1, 2, \ldots, N_q, \quad  r\in [-1, 1]^{|S||A|}\}.    
\end{equation}
The reward function $r$ is recovered within the intersection of two sets $\hat{\mathcal{R}}_\text{IRL}(\epsilon_g)$ and $\hat{\mathcal{R}}_\text{CHF}(\epsilon_r)$:
\begin{equation}\label{eq:irl-rlhf:c}
\hat{\mathcal{R}}_\text{IRL-CHF}(\epsilon_g, \epsilon_r) = \{ (r, u, v) \mid (u, v) \in \hat{\mathcal{R}}_\text{IRL}(\epsilon_g),\; r \in \hat{\mathcal{R}}_\text{CHF}(\epsilon_r), \; u = \hat{d}_e \circ r\}.
\end{equation}
It is important to note that $\hat{\mathcal{R}}_\text{IRL-CHF}(\epsilon_g, \epsilon_r)$ can become infeasible if $\epsilon_g$ or $\epsilon_r$ is set too small, due to the strict reward gap. Therefore, choosing proper values for $\epsilon_g$ and $\epsilon_r$ is crucial to ensure the feasibility of the LP.
The generalization guarantee provided in Theorem~\ref{thm:rlhf} also holds with $\mathcal{L}'$, under a similar proof.

\section{Detailed Experimental Setup}\label{app:experiment}

\begin{table}[htbp]
  \centering
  \caption{Algorithm Details}
  \begin{tabular}{@{}llll@{}}
    \toprule
    Algorithms & Description & Parameters \\
    \midrule
    MLE-IRL & Bi-level optimization algorithm for offline IRL~\citep{zeng2023understanding} &  Step size $=0.01$\\
    LP-IRL-1 & LP formulation of IRL~\eqref{eq:irl} with a moderate $\epsilon_g$ & $\epsilon_g = 0.01/\sqrt{N}$\\
    LP-IRL-2 & LP formulation of IRL~\eqref{eq:irl} with a tighter $\epsilon_g$  & $\epsilon_g  = 0.001/\sqrt{N}$ \\    
    LP-IRL-D & Integration of IRL and RLHF with discrete feedback~\eqref{eq:irl-rlhf:d} & $\epsilon_g=0.01/\sqrt{N}$, $\epsilon_r = 0.01/\sqrt{N}$\\
    LP-IRL-C & Integration of IRL and RLHF with continuous feedback~\eqref{eq:irl-rlhf:c}  & $\epsilon_g=0.1/\sqrt{N}$, $\epsilon_r = 0.01/\sqrt{N}$ \\
    \midrule
    MLE-HF & Pessimistic MLE under the BTL model~\citep{zhu2023principled} & $\lambda=0.1$, $B=1$\\
    LP-HF & LP formulation of RLHF~\eqref{eq:RLHFset} & $\epsilon_r = -0.01$\\
    \bottomrule
  \end{tabular}
\label{table:algorithms}
\end{table}

\paragraph{Environment setting and dataset.}
We consider an MDP with $|S|=10$, $|A|=2$, and $\gamma=0.95$.
The initial state distribution $\mu_0$ and state transition probabilities $P$ are randomly selected for each experiment.
Specifically, each element of $\mu_0$ and $P$ is generated from a uniform distribution in the range of $[0, 1]$, and then scaled to form probability distributions.
In each experimental run, we sample $N$ trajectories with a horizon of $H=20$.
$\pi_e$ is used for sampling trajectories in IRL, and the uniform policy ($\pi(a|s) = 1 / |A| \quad \forall (s,a) \in S \times A$) is employed for sampling queries (trajectory pairs) in RLHF.

\paragraph{Performance criteria.}
To introduce additional complexity to the problem, we have set the true rewards to have similar values: $r_{\text{true}}(s, a_1) = 1.0$ and $r_{\text{true}}(s, a_2) = 0.9$ for all states $s\in S$.
The performance of each algorithm is then assessed by measuring the proximity of an optimal occupancy measure under the true reward $r_\text{true}$ and the obtained reward function $\hat{r}$. 
Specifically, we report $\Vert d^*(r_\text{true}) - d^*(\hat{r})\Vert_1$, the $\mathcal{L}^1$ error between the true optimal occupancy measure $d^*(r_\text{true})$ and the optimal occupancy measure $d^*(\hat{r})$ under the estimated reward $\hat{r}$.
This error falls within the range of $0$ to $2$, with a value of $0$ indicating that the estimated optimal policy is equivalent to the true optimal policy. 
For each sample size $N$, we conducted 200 experiments and reported the mean and standard deviation of the error.

\paragraph{Expert setting.}
In offline IRL, the expert policy for sampling trajectories is set to $\pi_e = 0.52\times \pi^* + 0.48\times \pi_r$, where $\pi_r$ denotes a greedy policy that selects a suboptimal action.
This setting reflects that $\pi_e$ can deviate from $\pi^*$, particularly when all state-action pairs have similar rewards.
In offline RLHF, we consider two different types of human feedback: discrete feedback $y \in \{1, 2\}$ and continuous feedback $y \in [-1, 1]$. The discrete feedback is generated according to the BTL model under the reward $r_\text{true}$. The continuous feedback is generated from the uniform distribution, in the range between $0$ and $0.2 \times r^\top _\text{true}(\psi^1-\psi^2)$.

\paragraph{Algorithm details.}
Table~\ref{table:algorithms} provides a detailed description of the algorithms used in experiments. 
We employ a tabular setting for the reward function without any function approximation.
In LP-IRL-1 and LP-IRL-2, we assume that $2/3$ of trajectories are sampled from $\pi_e$, and the remaining samples are obtained from the uniform policy to estimate $\hat{d}_\text{sub}$.
In LP-IRL-D and LP-IRL-C, we assume that $2/3$ of trajectories are sampled from $\pi_e$, and the the remaining trajectories are sampled from the uniform policy to generate human feedback.
In all LP-IRL algorithms, we use the $\mathcal{L}^1$ norm constraint for $X$, and we set $\delta=0.1$ and $B=100$.
In MLE-HF and LP-HF, human feedback is given as discrete value following the greedy model.
In LP-HF, a reward function is selected from \eqref{eq:RLHFset} by optimizing a dummy objective function. 
For algorithm details of MLE-IRL and MLE-RLHF, please refer to \citep{zeng2023understanding} and \citep{zhu2023principled}, respectively.